# A Fuzzy Based Model to Identify Printed Sinhala Characters

G. I. Gunarathna, M. A. P. Chamikara
Department of Statistics and Computer Science
University of Peradeniya
Peradeniya, Sri Lanka
gihanigunarathna88@gmail.com, pathumchamikara@gmail.com

R. G. Ragel
Department of Computer Engineering
University of Peradeniya
Peradeniya, Sri Lanka
roshanr@pdn.ac.lk

*Abstract*— Character recognition techniques for printed documents are widely used for English language. However, the systems that are implemented to recognize Asian languages struggle to increase the accuracy of recognition. Among other Asian languages (such as Arabic, Tamil, Chinese), Sinhala characters are unique, mainly because they are round in shape. This unique feature makes it a challenge to extend the prevailing techniques to improve recognition of Sinhala characters. Therefore, a little attention has been given to improve the accuracy of Sinhala character recognition. A novel method, which makes use of this unique feature, could be advantageous over other methods. This paper describes the use of a fuzzy inference system to recognize Sinhala characters. Feature extraction is mainly focused on distance and intersection measurements in different directions from the center of the letter making use of the round shape of characters. The results showed an overall accuracy of 90.7% for 140 instances of letters tested, much better than similar systems.

## I. Introduction

As the printing process evolve, a need to make changes to already printed documents increased. Even to make minor changes, we have to retype the complete document and make necessary changes. In addition, the need to examine ancient books and word searching has increased.

As a result, techniques to digitize printed documents were explored. They allow users to easily modify and reprint the documents and also make word searching easier. Digitizing printed documents eliminates the need for retyping already printed documents for editing. It also allows users to make multiple copies of the printed document with clarity, different formatting, etc.

Optical Character Recognition (OCR) is the process of converting the image obtained by scanning a text or a document into a machine-editable format. This technique is mostly used to digitize printed documents, which involve scanning the printed document and using the resultant image to recognize characters. The scanned image is used to extract the features of characters. The recognition of characters was carried out by models designed using methods such as neural networks and fuzzy logic with the help of features extracted from the characters.

There are many digitizing solutions available for international languages such as English, which are already in wide use. However, converting printed Sinhala documents to editable text is not explored to the fullest and it is a challenge. The Sinhala language traces its origin to the Brahmi script of India. South Asian scripts such as Devanagiri, Telugu, Bengali, Tamil and Sinhala are descendants of this ancient Brahmi language. Sinhala language consists of 60 basic letters. These characters are round in shape and it is difficult to distinguish between characters that have minor variations. This unique feature of characters makes it a challenge to digitize them accurately.

A system, which makes use of these unique features, will overcome this challenge. The features which will be extracted from the image of the character will determine the efficiency and the accuracy of character recognition. The measurements from the center of a circular object will determine the variations of the circular shape. Therefore, the distance from the center of a character to the edges and number of intersections in several directions can be used as suitable features to extract in Sinhala character recognition.

Fuzzy logic [8] is the logic of approximation. The observed patterns of distance and intersection measurements with the help of fuzzy logic can be used to uniquely identify Sinhala characters.

This paper mainly concentrates on a novel OCR method to improve the accuracy of Sinhala character recognition using a fuzzy based model. The rest of the paper is organized as follows. Section II discusses the related work. The methodology followed on this research is discussed in Section III followed by the experimental set up and results in Section IV. Section V discusses the performance of the proposed system and the paper is concluded in Section VI.

## II. Related Work

OCR techniques are widely used for recognition of characters of many languages. Various researches have been conducted under OCR techniques followed by extracting different features and feeding them to different models for recognition. Many systems have been built for Latin languages like English, and Asian languages like Arabic [2, 3], Devanagiri [7], etc.

Among Asian languages, Tamil characters have certain similar features to Sinhala characters. Both Tamil [6] and Sinhala [10], [11] characters can be segmented in to three main segments as shown in Fig. 1.

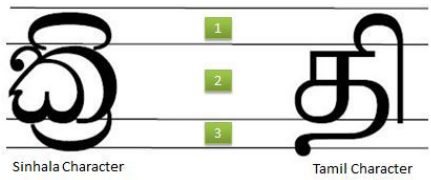

Fig. 1. A similarity between Sinhala and Tamil characters

Ramakrishnan and Mahata [6] have applied skew detection and correction in their multi font, multi size OCR application followed by Tamil character segmentation. The segmented characters are fed into a classifier for recognition. The classification strategy, first identifies the individual symbols, and in a subsequent stage, combines the appropriate number of successive symbols to detect the character. They have identified three segments that any Tamil character can be divided into and defined a three-level, tree structures classifier. This method has segmentation complexity.

Subramanian and Kubendran [9] have proposed an OCR application for Tamil printed characters defining a simple algorithm in preprocessing stage for skeletonization. A six dimensional feature space along with a similar method to the nearest-neighbor classifier has been used to recognize the characters. They have further stated that this approach also improves the accuracy of character recognition. However, it can be further improved by increasing the number of feature dimensions. It can be deducted from Subramanian's OCR system that the recognition power could be improved as the number of features extracted increase.

Different approaches to identify Sinhala characters were explored to observe unique features of Sinhala characters.

Premaratne and Bigun [4] have proposed a linear symmetric approach for printed Sinhala character recognition. The orientation features are used to recognize characters directly using a standard alphabet as the basis without the need for segmentation in to basic components. They have explored the unique features of Sinhala characters in the process of recognition of characters. Though this approach does not have segmentation complexity, many iterative filtering should be performed in order to recognize confusing characters.

In the OCR approach for Sinhala character recognition which has been proposed by Ajward et al. [1] preprocessing stages follows a neural network classifier. Extracted features such as centroid, bounding box are used to crop each individual image rather than feeding the characters directly to the classifier.

Rajapakse et al. [5] have proposed a neural network based character recognition system for Sinhala script with hidden layers to recognize hand-written Sinhala characters. In particular, it applies the widely used pattern classification neural network technique, the back propagation, in the recognition process. The segmentation method, which is used to partition the letters into 9 segments, was used to group the common features of Sinhala characters and extracting important features to help distinguish similar characters.

Therefore, there is a need of a novel method to recognize characters handling features of Sinhala language distinctly without the segmentation complexity. The feature set will be used to extract the relevant information from the input images in order to recognize, instead of directly feeding full sized images to the recognition model. The proposed solution in this paper concentrates on extracting suitable features from the Sinhala characters in order to increase the accuracy of recognition of characters.

III. METHODOLOGY

The proposed system measures two features of the letters and uses fuzzy logic for recognition. The system follows the six processes given, to recognize the characters.

1. Preprocessing
2. Distance measurement and calculation
3. Fuzzy Inference System 1 (FIS-1)
4. Intersection calculation
5. Fuzzy Inference System 2 (FIS-2)
6. Results and output

The complete design of the model which contains two fuzzy inference systems is shown in Fig. 2.

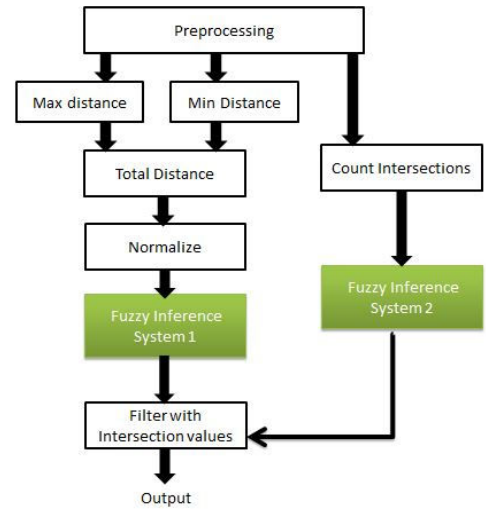

Fig. 2. System design

*A. Preprocessing*

As shown in Fig. 3, after converting the image into a binary image the preprocessing steps involve opening to reduce noise and closing to fill the holes in letters. Then the letter was cropped from the boundaries and skeletonized to obtain the real shape of the letter. Then spur was reduced and resized to a ratio of 7:5 in height and width. Finally, dilation was performed to fill the gaps created in letters after resizing.

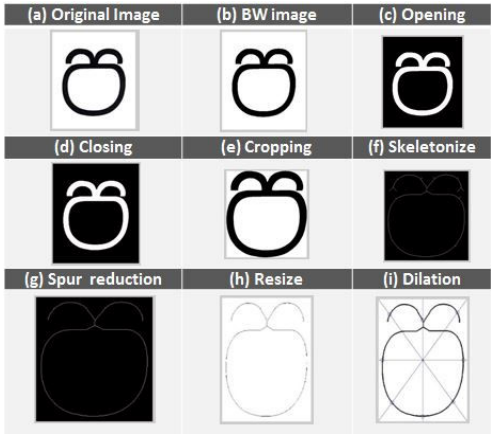

Fig. 3. Preprocessing stages (b) Binary image (c) Image after performing opening technique to reduce noise (d) Image after performing closing technique to remove holes (e) Cropped image (f) Skeletonized image (g) The image after spur reduction (h) Resized image (ratio of 7:5 in height and width) (h) Image after performing dilation to fill the gaps produced due to resizing.

### B. Distance measurement and calculations

The first feature extracted from the character is the distance from the center to the edges of the letter. Distance was measured along eight directions (W, E, N, S, NW, SE, SW and NE) as shown in the Fig. 4.

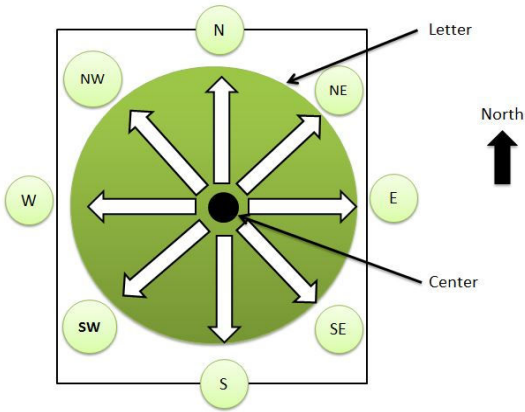

Fig. 4. Distance measurement in eight directions

Two measurements were taken along each direction as minimum distance ($D_{minimum}$) and maximum distance ($D_{maximum}$) for intersections (Fig. 5).

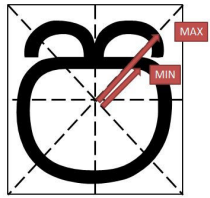

Fig. 5. Obtaining the minimum and maximum distances

Each value was normalized to a value between 0 and 10. Then the sum ($D_{Total}$) of minimum and maximum distances of one direction which varies between 0 and 20, was taken to obtain a significant output pattern as shown in Fig. 6.

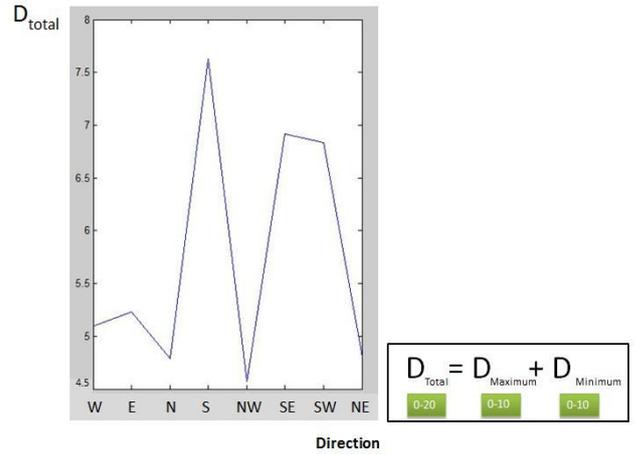

Fig. 6. Calculation and plot of the total distance

### C. Fuzzy Inference System 1

The Fuzzy Inference Systems (FIS) maps the fuzzy inputs provided with the fuzzy output based on the rule base and the input/output membership functions defined. This model uses Mamdani FIS to recognize characters. The FIS for distance with 8 inputs and one output is shown in Fig. 7.

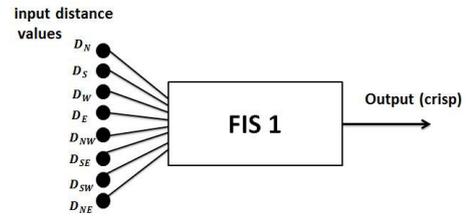

Fig. 7. FIS-1

The total distances calculated for eight directions were given as inputs to the FIS. Each input was categorized into one of 3 feature levels: low, medium and high. The three feature levels were represented in FIS as three membership functions for each input as shown in Fig. 8.

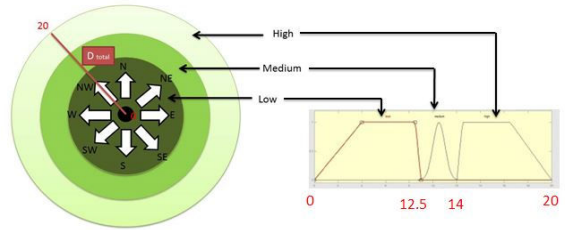

Fig. 8. Feature levels transferred to membership functions in FIS-1

Seven output membership functions were defined initially to recognize 7 basic types of Sinhala characters as shown in Fig. 9.

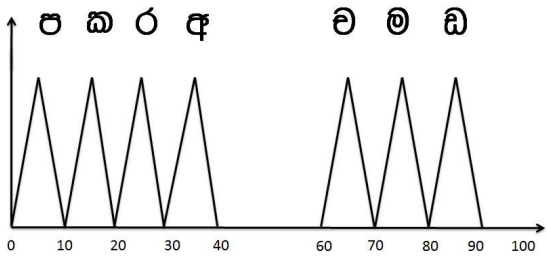

Fig. 9. Output membership functions of FIS-1

The corresponding crisp values were generated according to the recognized letter based on the initial rule base defined. Fig. 10 depicts how the rules were generated for characters.

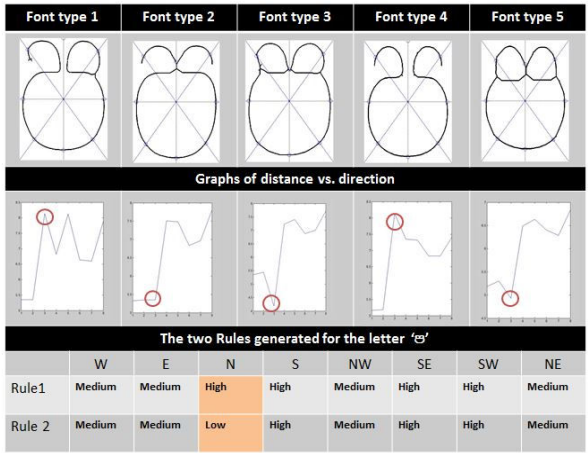

Fig. 10. Generation of rules for the letter 'm' in FIS 1

Five different font types of a character were observed as the training set to define the rules. As an example, Fig. 10 shows how two significant patterns were observed for the letter 'm'. First, the minimum and the maximum distances to the edges were measured along eight directions. Then, the graphs of distance vs direction were plotted to observe the significant patterns. The five patterns corresponding to the five font types vary only in the distance value of the direction 'N', as highlighted in the graphs of Fig. 10. Therefore, two rules were generated to identify the character 'm' in FIS-1.

The results obtained from FIS-1 could be improved since similar patterns were observed for different letters as depicted in Fig 11.

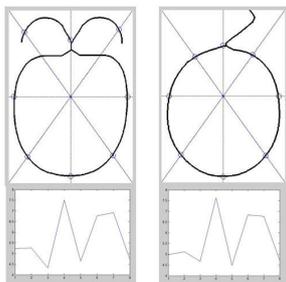

Fig. 11. Observed similar patterns of distance for different letters ('m' and 'r')

Therefore, these results were filtered with FIS-2 to distinguish between letters which have similar types of patterns for distance measurements.

*D. Intersection calculation*

The second feature extracted from the letter is the number of intersections from the center of the letter to the eight directions considered, which is depicted in Fig. 12.

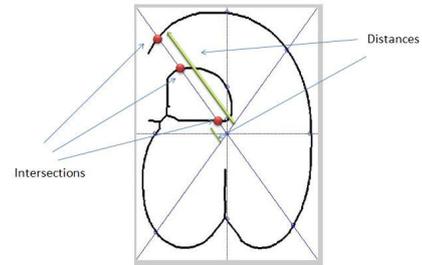

Fig. 12. Intersection measurement

*E. Fuzzy Inference System 2*

The numbers of intersections along each direction were taken as inputs to the FIS 2. The Mamdani FIS for intersections with 8 inputs and one output is shown in Fig. 13.

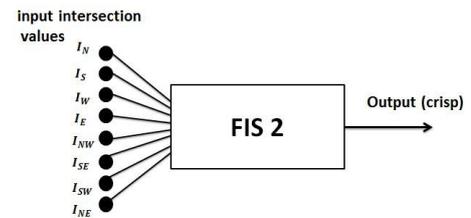

Fig. 13. FIS-2

Each input was categorized into one of 4 feature levels (low, low-medium, medium and high) since the number of intersections varies within a small range from 0 to 5. The four feature levels were represented in FIS as four membership functions for each input as shown in Fig. 14.

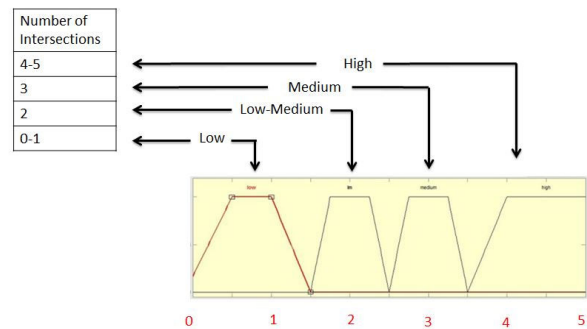

Fig. 14. The number of intersections along each direction transferred to input membership functions of each input of FIS-2.

As in FIS-1, the corresponding crisp values were generated according the rule base defined. Rules were generated for each letter observing the patterns obtained from 7 different letters in

5 font types to feed FIS-2. Fig. 15 depicts how the rules were generated for the characters.

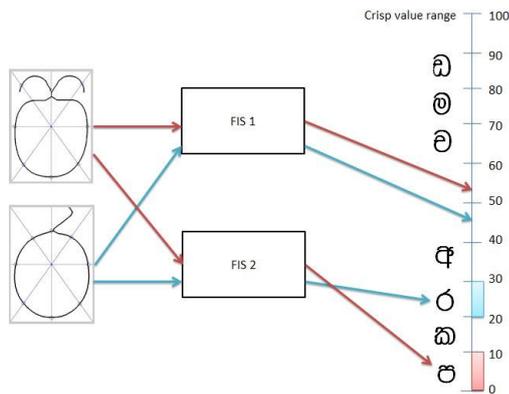

Fig. 15. Generation of rules for the letter 'm' in FIS 2

The training set of FIS-2 consists of five different font types from each letter. Fig. 15 highlights how five patterns of intersections corresponding to the five font types vary only in the intersection values of direction 'N'. 0 and 1 intersection values fall to the feature level 'low' while intersection value '2' falls to the feature level 'low-medium'. Therefore two rules were generated to identify the character 'm' in FIS 2 as shown in Fig. 15.

### F. Output

Finally the results of FIS-1 were filtered with the results of FIS-2 to increase the accuracy of the recognition of Sinhala characters. As depicted in Fig.16, the letter was recognized upon the optimal crisp value returned from either FIS-1 or FIS-2.

Fig. 16. Filtering of the results of FIS-1 with the results of FIS-2 to recognize characters 'm' and 'r'.

## IV. EXPERIMENTAL SETUP AND RESUTLS

After generating the rule bases for the two FISs using training set of 7 characters in 5 different font types, the system was tested under three test sets as shown in Table I. One character appeared only once in each OCR image of the testing sets.

TABLE I. THE TABLE OF CHARACTERS TESTED IN EACH STAGE

| Test set | Category of characters | | | | | | | |
|---|---|---|---|---|---|---|---|---|
| | 1 | 2 | 3 | 4 | 5 | 6 | 7 | 8 |
| 1 | | m | v | u | r | p | w | l |
| 2 | o | | i | g | | t | | y |
| 3 | b | | h | | | | | |

### A. Testing set 1

The purpose of testing the system in this character set was to check whether the system can easily distinguish between the basic categories of Sinhala characters. Therefore, 7 characters from 7 categories were selected to test the system with 20 different font types of each character including the training sets. Table II shows the results.

TABLE II. THE TABLE OF THE ACCURACY OF RECOGNITION IN FIRST STAGE

| Number of characters | Accuracy of characters in stage 1 | | | | | | |
|---|---|---|---|---|---|---|---|
| | m | v | u | r | p | w | l |
| Tested | 20 | 20 | 20 | 20 | 20 | 20 | 20 |
| Correct | 19 | 19 | 19 | 20 | 10 | 20 | 20 |
| Accuracy (%) | 95 | 95 | 95 | 100 | 50 | 100 | 100 |
| FIS 1 rules (52) | 2 | 2 | 16 | 7 | 2 | 8 | 15 |
| FIS 2 rules (18) | 2 | 3 | 4 | 1 | 2 | 3 | 3 |

The system gives an overall accuracy of 90.7% for the 140 instances of basic letters tested. This accuracy is greater than the overall accuracy of the printed Sinhala character recognition systems proposed by Premaratne and Bigun (84%), Ajward et al. (75%) and handwritten character recognition system proposed by Rajapakse et al. (75%).

### B. Testing set 2 & 3

The system was further improved to recognize more characters introducing another 7 characters of 5 font types while testing set 1 characters exist in the system. The purpose of this was to test whether the system can distinguish between similar types of characters (Table I) which fall in to the same categories. The rule base was introduced with more rules. The system distinguished between the characters successfully providing an accuracy of 98.2% for 5 font types of each letter while all letters from testing sets were in the system.

### C. Elapsed time

The time taken to recognize each character was measured and presented in Table III. The system returned an average time 1.38 seconds to recognize one character.

TABLE III. THE TIME ELAPSED FOR RECOGNITION OF CHARACTERS

| Font type | Time taken to recognize characters (seconds) | | | | | | |
|---|---|---|---|---|---|---|---|
| | *m* | *v* | *u* | *r* | *p* | *w* | *l* |
| *1* | 1.67 | 1.31 | 1.34 | 1.38 | 1.42 | 1.54 | 2.52 |
| *2* | 1.31 | 1.24 | 1.19 | 1.26 | 1.41 | 1.48 | 2.09 |
| *3* | 1.49 | 1.12 | 1.21 | 1.34 | 1.42 | 1.49 | 1.59 |
| *4* | 1.15 | 1.40 | 1.35 | 1.44 | 1.47 | 1.59 | 1.69 |
| *5* | 1.12 | 1.45 | 1.40 | 1.46 | 1.58 | 1.50 | 1.77 |
| *6* | 1.37 | 0.96 | 0.98 | 1.01 | 1.20 | 1.34 | 1.36 |
| *7* | 1.77 | 1.27 | 1.28 | 1.30 | 1.30 | 1.44 | 1.62 |
| *8* | 1.40 | 1.05 | 0.98 | 1.01 | 1.30 | 1.40 | 0.87 |
| *9* | 1.31 | 1.05 | 1.03 | 0.90 | 1.25 | 1.39 | 0.84 |
| *10* | 1.75 | 1.00 | 1.14 | 1.06 | 2.32 | 1.38 | 1.59 |
| *Varience* | 0.06 | 0.09 | 0.06 | 0.07 | 0.11 | 0.01 | 0.31 |

Fig.17 shows the graph of variances of elapsed time for the testing character set. Though the character sets such as 'l', 'p' and 'v' show high variances comparatively, the variance values still lie in a very low range proving that the algorithm performs efficiently in all situations.

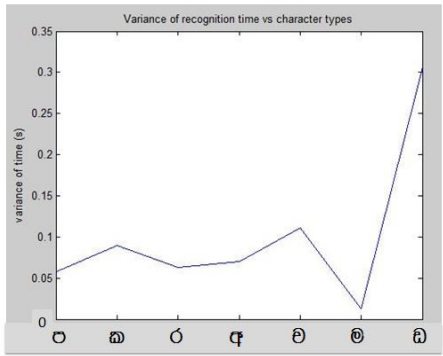

Fig. 17. Variance of elapsed time for 7 characters

## V. DISCUSSION

Identification of patterns in the distances and intersections to define the rule base was one of the important aspects of our system. The distance results filtered with intersection results improves the accuracy of recognition of Sinhala characters. Since, the system does not have segmentation complexity, the time taken to recognize a character is very short. The system always extracts the same number of features from all the letters. Therefore, overall accuracy is improved.

For an accuracy of 90.7% for a testing dataset of 140 characters prove that the proposed algorithm does not show any overfitting anomalies.

The system has only considered the maximum and minimum distances. The intersected distances in-between the maximum and the minimum were ignored. Therefore, the system can be improved by measuring these to increase its accuracy of recognition.

This system can be further improved by increasing the number of directions considered and introducing more complex characters to the system.

Further, the system can be introduced with another parameter as the sum of intersections of a letter in all directions to increase the accuracy of the recognition. Furthermore, the system can be optimized applying skew detection and correction technique to recognize the handwritten characters.

## VI. CONCLUSION

The main goal of this research was to provide an easy, efficient and correct system to improve Sinhala character recognition of the printed Sinhala documents.

The distance calculation from the center of the letter is proved to be an effective method to recognize characters after filtering the number of intersections in each direction. The recognition accuracy was improved compared to other similar systems in the literature.

Even though the system performs well in recognition for basic types of characters, it could be further improved to handle complex Sinhala characters.